\documentclass{article}
\usepackage{graphicx} % Required for inserting images
\usepackage[english]{babel}
\usepackage[letterpaper,top=2cm,bottom=2cm,left=3cm,right=3cm,marginparwidth=1.75cm]{geometry}
\usepackage{amsmath}
\usepackage{amsfonts}
\usepackage{mathtools}
\usepackage[colorlinks=true, allcolors=blue]{hyperref}
\usepackage{caption}
\usepackage{subcaption}
\usepackage{float}
\newcommand\norm[1]{\lVert#1\rVert}
\graphicspath{{figs/}} % tell it where figures live; here, in the subdir called "FIGS"

\title{neuralODE submission}
\author{bingxianxu2026 }
\date{November 2023}

\begin{document}

\begin{centering}
    \textbf{\Large Constructing interpretable principal curve using Neural ODEs}\\[3mm]
    \textbf{Guangzheng Zhang$^{1}$, Bingxian Xu$^{2,3*}$}\\[1mm]
    \textsuperscript{1}Guangming Laboratory, Shenzhen, China \\
    \textsuperscript{2}Department of Molecular Biosciences, Northwestern University, Evanston, IL 60208, USA;\\ 
    \textsuperscript{3}NSF-Simons Center for Quantitative Biology, Northwestern
University, Evanston, IL 60208, USA;\\
    $^{*}$To whom correspondence should be addressed. Email: bingxianxu2026@u.northwestern.edu.\\
\end{centering}

\begin{abstract}
The study of high dimensional data sets often rely on their low dimensional projections that preserve the local geometry of the original space. While numerous methods have been developed to summarize this space as variations of tree-like structures, they are usually non-parametric and ``static'' in nature. As data may come from systems that are dynamical such as a differentiating cell, a static, non-parametric characterization of the space may not be the most appropriate. Here, we developed a framework, the principal flow, that is capable of characterizing the space in a dynamical manner. The principal flow, defined using neural ODEs, directs motion of a particle through the space, where the trajectory of the particle resembles the principal curve of the dataset. We illustrate that our framework can be used to characterize shapes of various complexities, and is flexible to incorporate summaries of relaxation dynamics. \end{abstract}

\section{Introduction}
Much of modern biology is devoted into the understanding of cell state transitions and learning how these transitions occurs has led to discoveries such as induced plouripotent stem cells~\cite{takahashi_induction_2006}. Although the advancement of technologies have accelerated the pace at which new discoveries were made, there is a shortage of framework capable of providing insights. 

One of the most widely used method to generate mechanistic model in biology is through the use of ordinary differential equations (ODEs), where the rate of change of each measured feature such as protein or mRNA abundance is explicitly modeled as $\frac{dx_i}{dt}=f_i(x_1,x_2,\hdots, x_n)$, where $f_i$ represents reactions that are taking place. In perhaps all cases, more than one systems of ODE will be able to recapitulate experimental observation hence these models need to be constrained by prior information of biological knowledge such as gene regulatory networks or known biochemical kinetics. 

ODE models can be used to explain and predict experimental systems. For example, it has been used to understand the yeast GAL4 system, one capable of undergoing a bifurcation from monostability to bistability as its source of nutrient changes. By measuring the time series of GAL4 fluorescence, a model was trained to construct a bifurcation diagram and predict the presence of monostability/bistability given concentrations of available nutrients~\cite{venturelli_population_2015}. On the other hand, ODE model was used to understand the JAK-STAT signaling pathway~\cite{mudla_cell-cycle-gated_2020}, which was composed of a positive feedback loop mediated by STAT1 and a negative feedback loop mediated by USP18. By optimizing a simple ODE model, it was able to predict the presence of a delayed negative feedback loop, which was found to stem from the cell cycle. On top of understanding existing systems, ODE models have even been used to design synthetic genetic circuits. Starting from different ODE models constructed for different real oscillators, it was observed that the general requirement for biochemical oscillations were: negative feedback, time delay, nonlinearity and balancing of time scales~\cite{novak_design_2008}. These insights were applied to construct synthetic biochemical oscillators~\cite{hasty_synthetic_2002, stricker_fast_2008}, some even capable of prolonging lifespan~\cite{zhou_engineering_2023}. 

While mechanistic ODE models have accomplished much, there are problems that has led to other, more abstract ways to construct ODE models. For example, even for a simple biochemical process, there may be tens or even hundreds of different known proteins and mRNAs involved. While having a large number of interacting molecules can generate complex dynamical behaviours, it is often impossible to measure all participating species, and even for the measurable ones, their dynamics are usually quite simple, unable to sufficiently constrain the parameters to be fitted. To overcome such problems, it was first proposed to combine certain reactions to reduce the number of system parameters. One way of doing so is to adopt the quasi-steady state assumption (QSSA). QSSA states that if there present a separation of time scale between reactions, one can assume that the fast reactions are always at steady states thereby reducing the dimension of the system. While some need to reduce the dimension of systems constructed based on prior biological information, others may need to do the opposite due a complete lack of prior knowledge. Methods such as the Sparse Identification of Nonlinear DYnamics (SINDy)~\cite{brunton_discovering_2016} were developed for this exact purpose. It first constructs a library of possible terms and subsequently conduct model fitting multiple times to iteratively remove terms that do not contribute significantly to the measured dynamics. However, SINDy requries measurement of all involved species in order to reconstruct the full system. In situation where not all species involved are measurable, one can first guess the dimensionality of the system using Takkens' theorm~\cite{packard_geometry_1980} and then apply other methods designed to ``guess'' the dynamics of unmeasured variables~\cite{ribera_model_2022, daniels_automated_2015, daniels_automated_2019, ji_autonomous_2021, somacal_uncovering_2022}. However, even when one construct a model, capable of recapitulating experimental findings, it may still be too complicated to conduct any formal analysis.

% mention waddington's landscape
While the concept of Waddington's landscape was not invented to solve this very problem, the idea of envisioning cell state transitions as trajectories of balls rolling along a bifurcating landscape enabled a simpler parameterization of the dynamics. In addition, as motion on the landscape is always downhill, the landscape parameterization represents a natural hierarchical structure which is common in biological systems~\cite{saez_statistically_2022}. When the landscape $V(\vec{x})$ is defined, the dynamics of the system along each dimension can be easily generated as: $\frac{d\vec{x}}{dt} = -\frac{dV(\vec{x})}{d\vec{x}}$. This metaphorical landscape can be used to make predictions which can then be tested experimentally. For example, landscapes have been constructed from cell type labels and was used to understand and predict how pluripotent stem cells adopt to different terminal state as a function of different signaling factors~\cite{saez_statistically_2022}. However, this presents a problem as it is difficult to reliably assign cell types or to sample cells without bias. Another way to exploit the landscape idea while circumventing the cell type assignment and sampling issue came from single cell transcriptomics, where the high-dimensional data is usually visualized by projecting cells onto a two-dimensional space~\cite{mcinnes_umap_2020, haghverdi_diffusion_2015, qiu_reversed_2017, tsne}. This two dimensional space can be thought of as a conglomeration of ``photos'' taken from above of balls rolling down the transcriptomic landscape. While the destructive nature of single cell sequencing prohibits the attainment of trajectories of each cell, having these ``photos'' allows one to see locations on the landscape that have been frequented. One can then ``connect'' clusters of cells using principal curves~\cite{hastie_principal_1989, street_slingshot_2018}, which is essentially a nonlinear version of the principal component, and order cells accordingly. This order, or sometimes termed the pseudotime, is used to construct ``time series'' of gene expression, which can subsequently be used to identify critical genes such as those that govern fate decision. 

To summarize, there are a few hindrance during the construction of a landscape based view. 
First, though it is simpler to work in a low dimensional space, as was done by Saez et al.~\cite{saez_statistically_2022}, the construction of this space is not trivial. In the work of Sae et al.~\cite{saez_statistically_2022}, the analysis were conducted in an arbitrary 2D space that has no connection to the actual gene expression measurement, which means that movement on their landscape is not interpretable. In the setting of single cell RNA-seq, the 2D space is usually created via dimensionality reduction algorithms such as UMAP or TSNE~\cite{mcinnes_umap_2020, tsne}. Though these methods are capable of creating a low dimensional representation on which movements can be interpreted, there are only two major ways to characterize this movement. The first main approach (the tree approach), which assume that there is a mean path shared by all cells, constructs a minimum spanning tree on which data can be projected and ordered~\cite{street_slingshot_2018, qiu_reversed_2017}. The second main approach~\cite{la_manno_rna_2018} makes use of information contained within mRNA splicing. By looking at the quantity of spliced and unspliced mRNA, these methods estimate the rate of change of gene expression thereby predicting the ``movement'' of each cell. While the first approach cannot be used to generate predictions, the second requires smoothing by nearest-neighbor pooling, a technique that may skew downstream analysis~\cite{gorin_rna_2022}.

Here, we develop a flexible, data-driven framework to characterize the low dimensional manifold that is based on neural ODE~\cite{chen_neural_2018}. We show on simulated data that we can generate time independent flows that characterize complex geometries similar to the tree approach and compute the direction of movement of all points on the phase plane just like RNA velocity. With our approach, we can easily ``extrapolate'' outside of the training data and therefore compute quantities such as the finite time Lyapnouv exponent~\cite{krishna_finite_2023}. 

\section{Methods}
\subsection{Neural ODE}
We used the neural ODE framework to model the velocity field~\cite{chen_neural_2018}. The motion of a particle in the field will be defined as $\frac{d\Vec{x}}{dt}=g(\Vec{x})$, where $\Vec{x}$ is the location of the particle, and $g$ is a neural network that takes in the coordinate of the particle as input and output its direction of motion. Additionally, we set $\norm{g(\Vec{x})}=1$ to generate constant speed flow. The neural network $g$ is trained by the adjoint sensitivity method using the torchdiffeq package in python. 

\subsection{Computing finite time Lyapunov exponent}
Given two points on a flow field initially $\delta Z(0)$ apart, the Lyapunov exponent, $\lambda$, is defined as:
\begin{equation*}
    \norm{\delta Z(t)}\approx e^{\lambda t} \norm{\delta Z(0)}
\end{equation*}

Essentially, the Lyapunov exponent is positive if a small perturbation results in a big difference in and negative vice versa. To estimate the Lyapunov exponent numerically, we need first define the flow map operator $\Phi_{t}^{t+T}:\mathbb{R}^n \to \mathbb{R}^n$ as:
\begin{equation*}
    \Phi_{t}^{t+T}:x(t) \to x(t) + \int_{t}^{t+T} g(x(\tau)) d\tau.
\end{equation*}

To compute FTLE~\cite{krishna_finite_2023}, we initialize by setting up a grid of points on our flow field indexed by $i$ and $j$. Then, we compute the flow map Jacobian, defined as:
\begin{align*}
        \left(\mathbf{D}\Phi\right)_{i,j} &= 
    \begin{bmatrix}
        \frac{\Delta x(T)}{\Delta x(0)} & \frac{\Delta x(T)}{\Delta y(0)}\\
        \frac{\Delta y(T)}{\Delta x(0)} & \frac{\Delta y(T)}{\Delta y(0)} \\
    \end{bmatrix} \\
    &= \begin{bmatrix}
        \frac{x_{i+1,j}(T) - x_{i-1,j}(T)}{x_{i+1,j}(0) - x_{i-1,j}(0)} & \frac{x_{i,j+1}(T) - x_{i,j-1}(T)}{y_{i,j+1}(0) - y_{i,j-1}(0)}\\
        \frac{y_{i+1,j}(T) - y_{i-1,j}(T)}{x_{i+1,j}(0) - x_{i-1,j}(0)} & \frac{y_{i,j+1}(T) - y_{i,j-1}(T)}{y_{i,j+1}(0) - y_{i,j-1}(0)} \\
    \end{bmatrix}
\end{align*}
With the flow map Jacobian, we compute the Cauchy-Green deformation tensor:
\begin{equation*}
    \Delta_{i,j} = \left( \mathbf{D}\Phi \right)^T \mathbf{D}\Phi
\end{equation*}
Lastly, the FTLE, $\sigma_{i,j}$, at point $(x_i,y_j)$ is defined as:
\begin{equation*}
    \sigma_{i,j} = \frac{1}{T} \ln{\sqrt{(\lambda_{max})_{i,j}}},
\end{equation*}
where $\lambda_{max}$ is the largest eigenvalue of $\Delta_{i,j}$.
FTLE computation was conducted using custom python code. 

\subsection{Phase response curve and the application of perturbation}
The phase response curve is simulated with the following parameter choices.
\begin{center}
\begin{tabular}{c|c}
    \hline
    parameter & value \\
    \hline
     $\sigma_\phi$ &  0.05\\
     $\xi_1$ & 0\\
     $\xi_2$ & 0\\
     $A_1$ & 0.4\\
     $A_2$ & 0.2
\end{tabular}
\end{center}
Additionally, the final PRC was scaled by 100 to cover a reasonable amount of the unit circle. To perturb the system, we instantaneously increase/decrease the amplitude of the oscillation while keeping its phase constant. 

\section{Results}
\subsection{Principal flow}
Denote data points as $\mathbf{X} \in \mathbb{R}^p$. A curve $f(\lambda)$, is said to be a principal curve if the projection of $\mathbf{X}$ onto $f(\lambda)$, in other words point on $f(\lambda)$ that is the closest to $\mathbf{X}$, denoted as $\lambda_f(x)$, satisfies:
\begin{equation*}
    E(\mathbf{X} \mid \lambda_f(\mathbf{X}) = \lambda) = f(\lambda).
\end{equation*}
This definition of the principal curve provides some intuition for how it should be constructed algorithmically. One can first project $\mathbf{X}$ onto the first principal component, and then for each data point, find all points whose projections are nearby and use their average projection as the updated $f(\lambda)$~\cite{hastie_principal_1989}. In this formulation of the principal curve, the curve itself is represented as n tuples ($\lambda_i, f_i$), where n is the sample size. In the context of analyzing data from biological systems, this representation of the data provides limited mechanistic insight and cannot be used to make predictions. To overcome this limitation, we constructed a velocity field $v(x,y)$ such that the flow within it can be reminiscent of a principal curve and we termed this flow the principal flow.  

\subsection{Constructing principal flow using Neural ODE}
The velocity field is constructed using neural ordinary differential equation~\cite{chen_neural_2018} for its ability to deal with data that comes with limited prior information. In our formulation, we define the velocity field as:
\begin{equation*}
    \frac{d\Vec{x}}{dt} = g(\Vec{x}),
\end{equation*}
where g is a neural network.  For simplicity purpose, we constrained flows on this velocity field to have constant velocity by enforcing $\norm{\frac{d\Vec{x}}{dt}}=1$. 

\begin{figure}[h]
\centering
\includegraphics[width=1\textwidth]{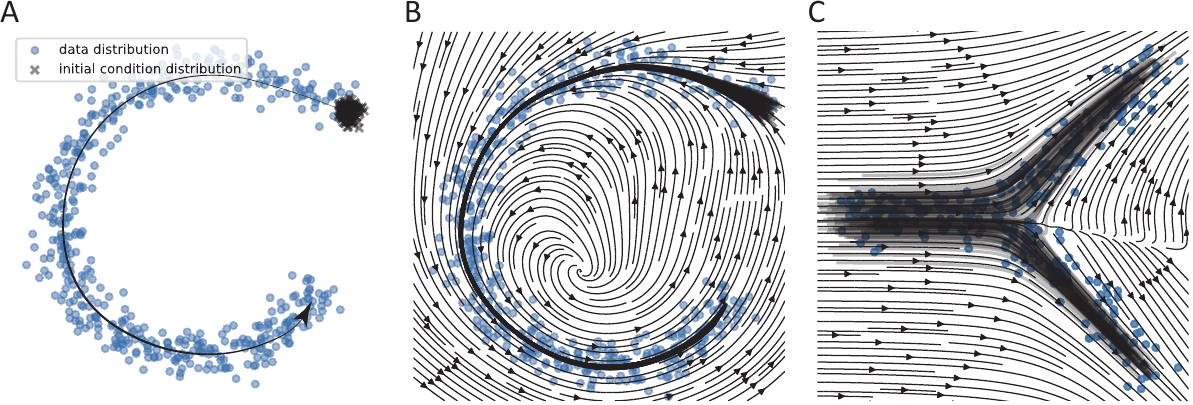}
\caption{A: Method outline. A velocity field that carries the initial condition through the observed data. Example velocity field generated for the ``C'' (B) and ``Y'' (C) distribution. Black curves represent trajectories within this velocity field.}
\label{fig:fig1}
\end{figure}

To train the neural network, the only input we require is the probability distribution, $p(\Vec{z})$ of the initial condition, $\Vec{z}(t=0)$ and the target data $\mathbf{X}$ (Figure~\ref{fig:fig1}A). The loss function is constructed as:
\begin{equation*}
    \mathcal{L}=\sum_{i,t} \norm{\min \left( \Vec{z_i}(t) - \Vec{x} \right )} + \sum_i \norm{\min \left( \Vec{z}(t) - \Vec{x_i}\right)},
\end{equation*}
where the first term ensures that the simulated trajectories are always close to the observed data, and the second enforces the simulated trajectories to travel through all observed data. 

\subsection{Principal flows of complex geometries}
To test whether this framework can recapitulate the observed data distribution, we first constructed a ``C'' shaped distribution for its non-linearity and lack of branches and found that our approach can successfully reconstruct a velocity field that directs flow through the observed data (Figure \ref{fig:fig1}B). Additionally, we observed the path taken by each ``particle'' in the flow field converges to form a principal curve.

Having illustrated that our algorithm can construct principal flow in the previous example, we sought to test its performance on more complex geometries. As it is common in data from biological systems to contain branching trajectories, we constructed a ``Y'' shaped distribution and observed that our method is capable of producing a velocity field that contains a separatrix, which defined the boundary of two branching trajectories (Figure~\ref{fig:fig1}C). In this example, the path taken by each particle do not converge because we have enforced our system to be time independent. As a time invariant flow filed cannot have crossing paths, it is not possible to generate a true principal curve as we have with the ``C'' distribution. 

\begin{figure}[h]
\centering
\includegraphics[width=1\textwidth]{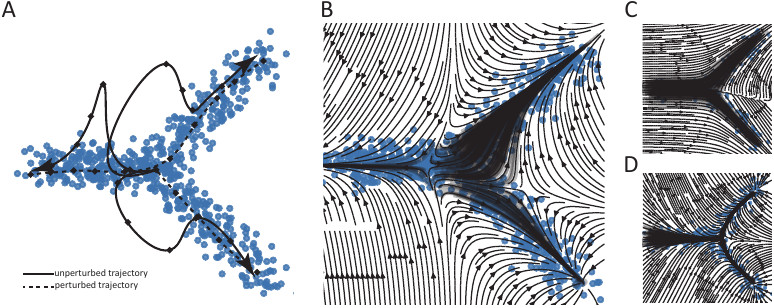}
\caption{A: Comparison between trajectories with and without noise injection. Noise injection forces the perturbed trajectory back to the data manifold, providing more attraction. B: Fitted velocity field generating three separate trajectories with the noise injection procedure. C: ``Y'' distribution with two branches without noise injection. D: ``Y'' distribution with two branches with noise injection.}
\label{fig:fig2}
\end{figure}

Additionally, we tested if our method is capable of generating a flow where points were initialized at the ``center'' of the ``Y'' shape that subsequently branch off into three separate trajectories and experienced no success using our previous approach. However, we found success with a slightly more complicated procedure where noise is injected at the end of each time step (Figure~\ref{fig:fig2}A, B). As this procedure appears to provide the data manifold with additional ``attraction'', we tested how this noise injection procedure effects the properties of reconstructed velocity field. Interestingly, we found that while we have constrained our flow to have unit speed, this procedure is able to generate fixed points, providing additional flexibility to our method (Figure~\ref{fig:fig2}C, D). 

\subsection{Finite time Lyapunov Exponent}
An important reason for the construction of the entire velocity field is its ability to predict the effect of a perturbation. This is highly relevant to biology in that it allows one to identify, for example, cell states that are the most susceptible to drug treatment. To pinpoint locations on the velocity field where a ``tiny'' perturbation will have the largest effect, we computed the finite time Lyapunov exponent (FTLE). 

\begin{figure}[h]
\centering
\includegraphics[width=1\textwidth]{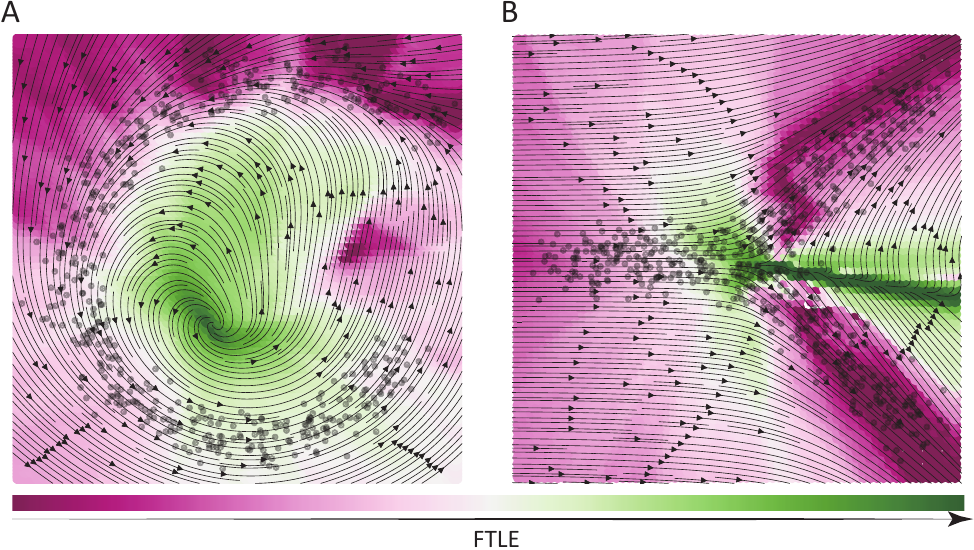}
\caption{FTLE field computed for the reconstructed flow field for the ``C'' (A) and ``Y'' (B) distribution.}
\label{fig:fig3}
\end{figure}

Given that we have constrained our flow with a constant velocity, all computed FTLEs are positive. Through the direct visualization of FTLE on the reconstructed velocity field, we were able to identify locations with small FTLE, which are insensitive to perturbation. In velocity fields where a separatrix is present, its surrounding area will have a large FTLE as a small perturbation nearby is sufficient to push the system from one branch to another (Figure~\ref{fig:fig3}A, B).

\subsection{Simple model of the human circadian rhythm}
% add things regarding perturbative experiments
In the previous cases where the loss function is defined as:
\begin{equation*}
    \mathcal{L}=\sum_{i,t} \norm{\min \left( \Vec{z_i}(t) - \Vec{x} \right )} + \sum_i \norm{\min \left( \Vec{z}(t) - \Vec{x_i}\right)},
\end{equation*}
the only constraint we have on the system, apart from being unit speed and time independent, is just the shape of the data distribution. As can be seen from figure~\ref{fig:fig2}C and D, the general feature of the reconstructed velocity field can drastically change even though both are capable of generating principal flows that spans the entirety of the data distribution. This suggests that passive observation is providing insufficient constraint and an accurate reconstruction of the underlying velocity field may also require some active effort.   
As we can easily incorporate additional term into our loss function, our approach can include measurements from perturbative experiments in the form of mean squared error. As an example, we tested whether our method can be used to reconstruct a velocity field using phase response curves (PRC), a commonly used measurement in circadian biology that captures how an oscillator respond to a small, precisely timed perturbation. Given that we would not have enough information to incorporate all the detailed molecular interactions that are present in the making of the circadian oscillation, motivated by the Kuramoto model where only phases of the oscillators are considered, we devised a representation of circadian oscillation by modeling a particle traveling along a unit circle, taking into account both phase and amplitude. When unperturbed, the system travels along the unit circle with unit speed. We choose to reproduce a type two PRC for its complexity, and it is defined as~\cite{huang_minimal_2022}:
\begin{equation*}
    M(\phi)=\sigma_\phi - A_1 \sin(\phi - \xi_1) - A_2 \sin(2\phi + \xi_2).
\end{equation*}
In this scenario, our loss is composed of two terms:
\begin{equation*}
    \mathcal{L}= \left (M(\phi) - \hat{M}(\phi) \right)^2 + \mathbf{E}[(x(t)^2 + y(t)^2)-1],
\end{equation*}
where the first term is the mean squared error between the observed and simulated phase response curves and the second terms ensures that the system travels on a unit circle. 

What is worth noting is that as we have constrained our flow to have unit speed, phase advance and phase delay must be associated with a transient decrease and increase in amplitude respectively and that the perturbation itself is represented as an instantaneous shift of oscillation amplitude. If this model is a somewhat adequate representation of the real circadian oscillation, it implies that phase shift can be induced by a transient manipulation of the oscillation amplitude. 

\begin{figure}[h]
\centering 
\includegraphics[width=1\textwidth]{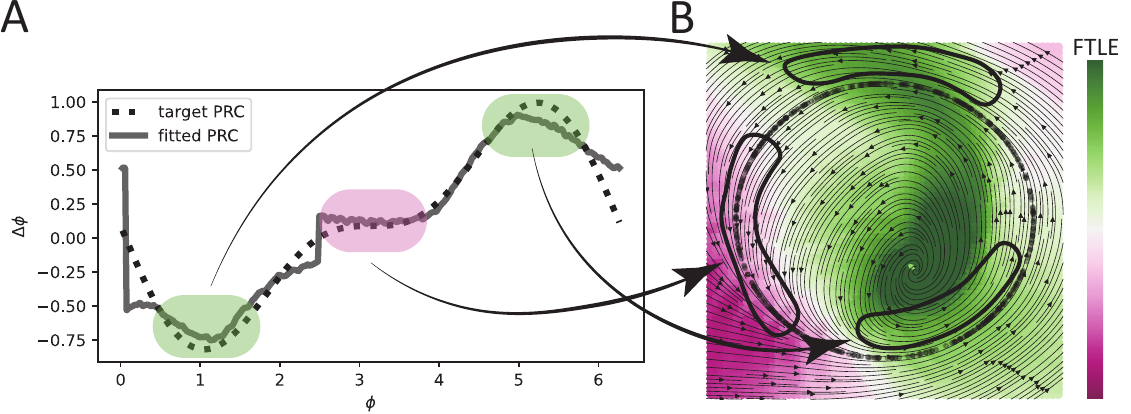}
\caption{A: Target and fitted phase response curve. Shaded area were associated with regions with high or low FTLE on the reconstructed velocity field (B)}
\label{fig:fig4}
\end{figure}

Interestingly, we observed that while we are able to construct a flow filed capable of producing a PRC that is similar to the objective (Figure~\ref{fig:fig4}A), our model of the circadian oscillation and the reality is very different in that our model cannot be perturbed without inducing any phase shift because any changes in the oscillation amplitude, no matter how small, will manifest as phase shift in the long term. By computing the FTLE of the reconstructed flow field, we found high FTLE near locations experiencing large phase shift when perturbed and low FTLE near resilient locations as expected (Figure~\ref{fig:fig4}B) 

\section{Discussion}
In this work, we have shown that neural ODE can be a powerful technique to construct flow fields that direct the traversal through observed data without any prior assumption. We demonstrate the capability of our method to create nonlinear and branching trajectories. In addition, using the circadian rhythm as an example, we demonstrate a workflow where the model is constructed from the combined input of measurements taken from a steady and perturbed state. 

Our method is a special form of time series modeling as we use ODEs to capture our system but do not use time explicitly in the fitting process. While seemingly absurd, data of this type is common in the field of single cell RNA-seq, where a cell can only produce one single measurement. Additionally, knowing time is not helpful when the systems of interest, such as cells maturing at different speed, are following the same trajectory in the phase space but with a different time scale. 

We do not constrain the system with a set of guessed equations~\cite{brunton_discovering_2016, ribera_model_2022}. Instead, our system is constrained by enforcing unit speed flow and time independence. To our knowledge, all existing method~\cite{pmlr-v119-tong20a, farrell_inferring_2022, hossain_biologically_2023} using neural ODEs to construct dynamical models have allowed the flow to be time dependent. We acknowledge that unit speed may not reflect the real system. However, our approach can easily incorporate temporal information when it is available. In addition, the fact that more than one neural networks will be able to generate the desired velocity field underscores the importance of having perturbative measurements to constrain the system further.   
  
We would like to argue that perturbative measurements or relaxation dynamics, are quintessential for developing a proper understanding of the system, and even more so in biology. As we have seen in our result, measured data only constrain the velocity field of its vicinity. In reality, this suggests that measurements taken from real experiments cannot provide any insight to predict how the system will behave when sufficiently perturbed. However, perturbing the system in a sufficiently large way is precisely what we want to achieve when we wish to reverse aging, remove cancer or regenerate tissues. Only when perturbative measurement is incorporated in a flexible framework like ours, can mathematical models finally be used to carry out the role they are designed to play: predict and control. 

\clearpage
\bibliographystyle{unsrt}
\bibliography{reference}
\end{document}